\documentclass[preprint,12pt]{elsarticle}

\usepackage{amssymb}
\usepackage{longtable}
\usepackage{amsmath}
\usepackage{subfig}
\usepackage{float}
\usepackage{multirow}
\usepackage{booktabs,arydshln}
\usepackage{bm}
\usepackage{pdflscape}
\usepackage{url}

\makeatletter
\def\adl@drawiv#1#2#3{%
        \hskip.5\tabcolsep
        \xleaders#3{#2.5\@tempdimb #1{1}#2.5\@tempdimb}%
                #2\z@ plus1fil minus1fil\relax
        \hskip.5\tabcolsep}
\newcommand{\cdashlinelr}[1]{%
    \noalign{\vskip\aboverulesep
           \global\let\@dashdrawstore\adl@draw
           \global\let\adl@draw\adl@drawiv}
  \cdashline{#1}
  \noalign{\global\let\adl@draw\@dashdrawstore
           \vskip\belowrulesep}}

\usepackage{lineno}

\journal{Applied Energy}

\begin{document}

\begin{frontmatter}



\title{Time to Market Reduction for Hydrogen Fuel Cell Stacks using Generative Adversarial Networks}


\author[inst1]{Nicolas Morizet}
\author[inst1]{Perceval Desforges}
\author[inst1]{Christophe Geissler}

\affiliation[inst1]{organization={Advestis},
            addressline={69 Boulevard Haussmann}, 
            city={Paris},
            postcode={75008}, 
            country={France}}

\author[inst2]{Elodie Pahon}
\author[inst2]{Samir Jemei}
\author[inst2]{Daniel Hissel}

\affiliation[inst2]{organization={FEMTO-ST Institute},
            addressline={FCLAB, Univ. Bourgogne Franche-Comte, CNRS}, 
            city={Belfort},
            country={France}}

\begin{abstract}
To face the dependency on fossil fuels and limit carbon emissions, fuel cells are a very promising technology and appear to be a key candidate to tackle the increase of the energy demand and promote the energy transition. To meet future needs for both transport and stationary applications, the time to market of fuel cell stacks must be drastically reduced. Here, a new concept to shorten their development time by introducing a disruptive and high-efficiency data augmentation approach based on artificial intelligence is presented. Our results allow reducing the testing time before introducing a product on the market from a thousand to a few hours. The innovative concept proposed here can support engineering and research tasks during the fuel cell development process to achieve decreased development costs alongside a reduced time to market.
\end{abstract}




\end{frontmatter}



\section{Introduction}
\label{sec:intro}
In 2021, more than 20 countries had already published or were actively preparing their national strategies relating to hydrogen~\cite{IEA2021}. These strategies are commonly based on three pillars: hydrogen production~\cite{Serra2020,Clark2017,Glenk2019}, hydrogen storage and distribution~\cite{Gallagher2016,Gallagher2017,Zhang2017}, and hydrogen usages~\cite{xia2016}. Regarding the latter, hydrogen can be used directly in chemical industry and in combustion engines, but also converted to electricity (and heat) through fuel cells dedicated to stationary~\cite{Pellegrino:2015,Wei:2017} and transport applications~\cite{edito2016,Cullen2021,Zhang:2020,Kim:2020}. The global fuel cell market has been sharply increasing in the last couple of years~\cite{Edito2021} considering the very large range of possible applications. In this dynamic framework and to meet the market requirements as well as possible, two main research areas are still strongly investigated from the industrial and academic point of views: the increase of efficiency and durability of fuel cell stacks and systems together with the reduction of their cost~\cite{Edito2021,Palencia:2017}. Obviously, in the V-model of a project aiming at developing a new fuel cell stack, testing and verification of both performances and durability of the developed product play a key role in terms of duration and costs. 
To reach these objectives, a disruptive solution involving generative adversarial networks (GAN) to decrease these fuel cells’ test duration and validation costs is developed. GAN have been already used for the real-time smart generation control of microgrids~\cite{Han:2022,Yin:2021,Dumas:2022,Dong:2022} as well as for the electricity prices forecasting~\cite{Lu:2022} even by including data augmentation for time series regression~\cite{Demir:2021}. It is also used for diagnostic tasks such as in~\cite{Li:2021},to detect failures on chiller systems to significantly reduce energy consumption and improve energy efficiency of buildings. GAN are also applied in the lithium-ion battery field for capacity estimation. GAN is utilized to obtain meaningful latent variables extracted from the impedance spectroscopy~\cite{Kim:2022}. However, this approach has not yet been applied to fuel cells. However, it seems quite appropriate method to contribute significantly reducing the time to market for a given new generation of fuel cell stack.


\section{Time to market for fuel cell systems - state of the art}
\label{sec:sota}
Without any reusable previous experiment, it is estimated that the development time for a new fuel cell system is 15 to 20 years~\cite{LSU:2022}. This very long development time includes all stages, from the selection of the best materials for the catalyst to the integration into a complete system, including the development of bipolar plates and their industrialization. When considering integration of existing fuel cell stacks into systems by OEM (Original Equipment Manufacturer), this development time can be reduced to about 18 to 24 months. Nevertheless, in both situations, functional and performance testing of the fuel cell stack is required, mostly done according to international test protocols~\cite{StackTest:2015}. These tests, even excluding long duration tests, are highly costly and time-consuming. A full characterization of a stack in various operating conditions, as defined in these international test protocols, can be estimated to a couple of weeks. 

Moreover, focusing on industrial production, and according to an analysis made for the US Department of Energy in 2017~\cite{Battelle:2017}, total testing time per produced fuel cell stack is assumed to be about 2.5 hours. This testing and validation time includes the connection of the stack to the appropriate air, hydrogen, and cooling subsystems and to an appropriate controlled load bank. This figure must also be put in parallel with the time required to manufacture a whole fuel cell stack, estimated to 13 minutes for an automotive fuel cell stack in 2021~\cite{IPA:2021}, leading to an end-of-line test time 10x greater than assembly time.

Once a stack is assembled and validated on the production line, the next step is to characterize its performance under different operating conditions, in order to qualify it for integration in various kind of power modules. The reference test is primarily the stack voltage obtained for different values of the drawn current (so-called polarization curves) under the applied test conditions. Polarization curves are often baseline measurements to qualify PEMFC (Proton Exchange Membrane Fuel Cell) stacks and components (e.g. catalyst, membrane, membrane electrode assembly, bipolar plates, etc) for given test conditions. To perform a polarization curve, a normalized test procedure has to be followed. It consists in different steps: start-up, conditioning, stabilization and measurements. Other recommended tests, also in different operating conditions, are the measurement of the voltages of the individual cells, the coolant outlet temperature (considered as the stack temperature), the outlet temperatures of the reactants (hydrogen and air) and the inlet and outlet pressures of the reactants.

With a rapidly expanding market, recent studies are looking at ways to further reduce the time required to break-in a fuel cell which is currently not suitable for large scale industrial applications. The time spent on the conditioning bench and the associated costs, mainly based on hydrogen consumption, must also be reduced~\cite{Balogun:2020}.

\section{Results}
\label{sec:results}
We have built and tested a data generator capable of producing artificial data in a short time that offer a high degree of statistical fit to the results of real experiments.
These results come directly from the test bed, which produced data during five consecutive days of testing under normal operating conditions. The retained experimental design consists in the variation of 10 input variables (\textit{i.e.} Tout$_{cooling water}$, Tin$_{H_2}$, Tin$_{Air}$, Pin$_{Air}$, Pin$_{H2}$, Qin$_{Air}$, Qin$_{H_2}$, RHin$_{Air}$, RHin$_{H2}$, Q$_{cooling water}$) across value ranges considered valid by the experimenter. 52 physical measurements including the previous 10 input variables, the main output variable representing the stack output voltage, and 41 other output variables (Table~\ref{tab:data}) were recorded at a frequency of $\approx 1$Hz. In total, we have 30,901 observations corresponding to the stabilization and polarization steps of the fuel cell. Depending on the experiment, extra categorical features (\textit{day} and \textit{step name}) are added to the continuous ones.

\subsection{Training}
\label{subsec:training}
Model training is performed on stabilization and polarization data taken on a five days span (very short period compared to the expected lifetime of the fuel cell). Input continuous data are standardized and categorical variables are one-hot encoded before being fed to a Generative Adversarial Network (GAN). In this work, the data reproduced are not time series (an attempt to reproduce synthetic energy time series using recurrent GANs can be found in \cite{Fekri:2020}), but synchronous snapshots of all the variables of interest from the stabilization and polarization phases. The measurements do not show any time dependency - besides the particular schedule retained for the physical inputs (temperature, pressure, current) in the experiments. Therefore, the choice made for this experiment was to validate some physical, non-temporal relationships between variables (e.g cell output voltage and stream intensity). The training ends after 25,000 epochs (each epoch corresponds to one batch of training data used to update the parameters of both networks once), when the critic can no longer improve its ability to distinguish artificial data from real data (their critic scores are very close), and the generator can no longer improve the quality of artificial data sent to the critic. On a local machine with a NVIDIA GeForce GTX 1080 Ti equipped with 12GB of video memory, the full training takes $\approx1$ hour. The architecture and the hyperparameters of the network are detailed in the \textit{Methods} section.

\subsection{Inference}
\label{subsec:inference}
Once the training is finished, the GAN model is completely described by $\approx$ 218,000 parameters. The generator is then detached and saved as a stand-alone function that can be called at will. This function takes as input a vector of 150 coordinates corresponding to independent draws of normalized Gaussian variables called the latent space. It is often agreed \cite{Pinetz:2020} that increasing the dimension of the latent space decreases the distortion of data, at the cost of a polynomial increase in the computation time. The latent space is mapped onto the space of real variables by the non-linear generator function. The latter provides as output vectors of length 52 (as many entries as continuous variables to reproduce), which are supposed to mimic experimental data performed at different physical conditions. The generator function can be called on a machine with a much smaller configuration than the one used for training. This situation is typical of neural networks, which require large computational resources (RAM and GPU) in training phase, but much smaller resources in inference phase. For example, the detached generator can produce 100,000 vectors of 52 artificial data on a 12 cores CPU, in $671.12$ms $\pm 0.024$ (statistics obtained for 500 runs), while requiring at most 2 GB of RAM.

\subsection{Results qualification}
\label{subsec:res_qualif}
The quality assessment of data produced by a GAN has been mostly studied in the field of image generation \cite{Salimans:2016, Heusel:2018, Shmelkov:2018, Wang:2019}. Establishing the quality of artificial tabular data representing physical quantities requires different methods from those usually used for image processing. The qualitative role of the human eye cannot be directly transposed to the evaluation of physical tabular data. The verification of the closeness between the distributions of the real and generated data focuses on the quality of reproduction of the correlations between the variables. We introduce several useful metrics for data comparison.

\subsection{Correlation matrices}
\label{subsec:corr_mat}
\begin{figure}[htp!]
    \centering
    \subfloat[\centering Real data.]{{\includegraphics[width=0.5\textwidth]{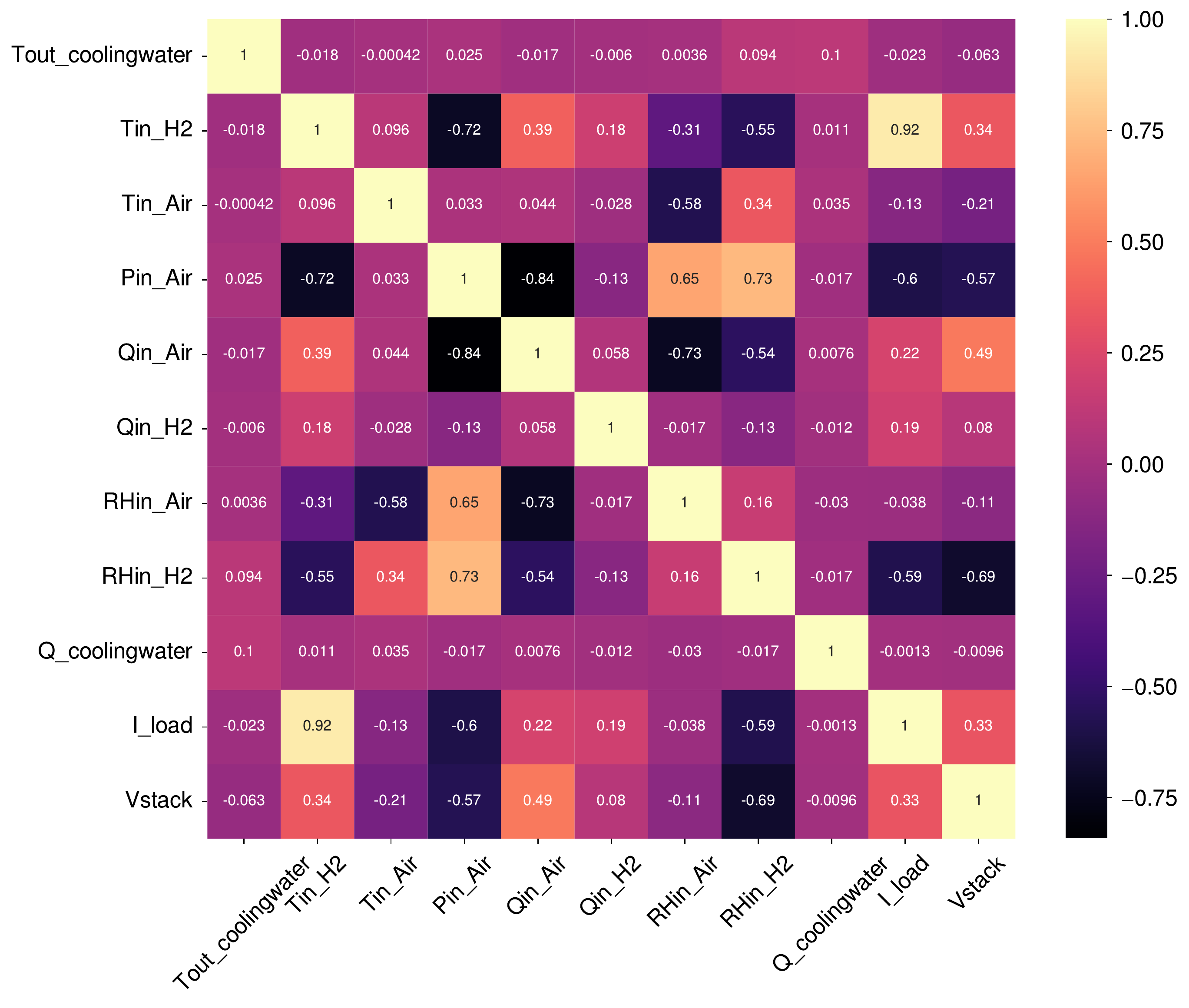} }}
    \subfloat[\centering Generated data.]{{\includegraphics[width=0.5\textwidth]{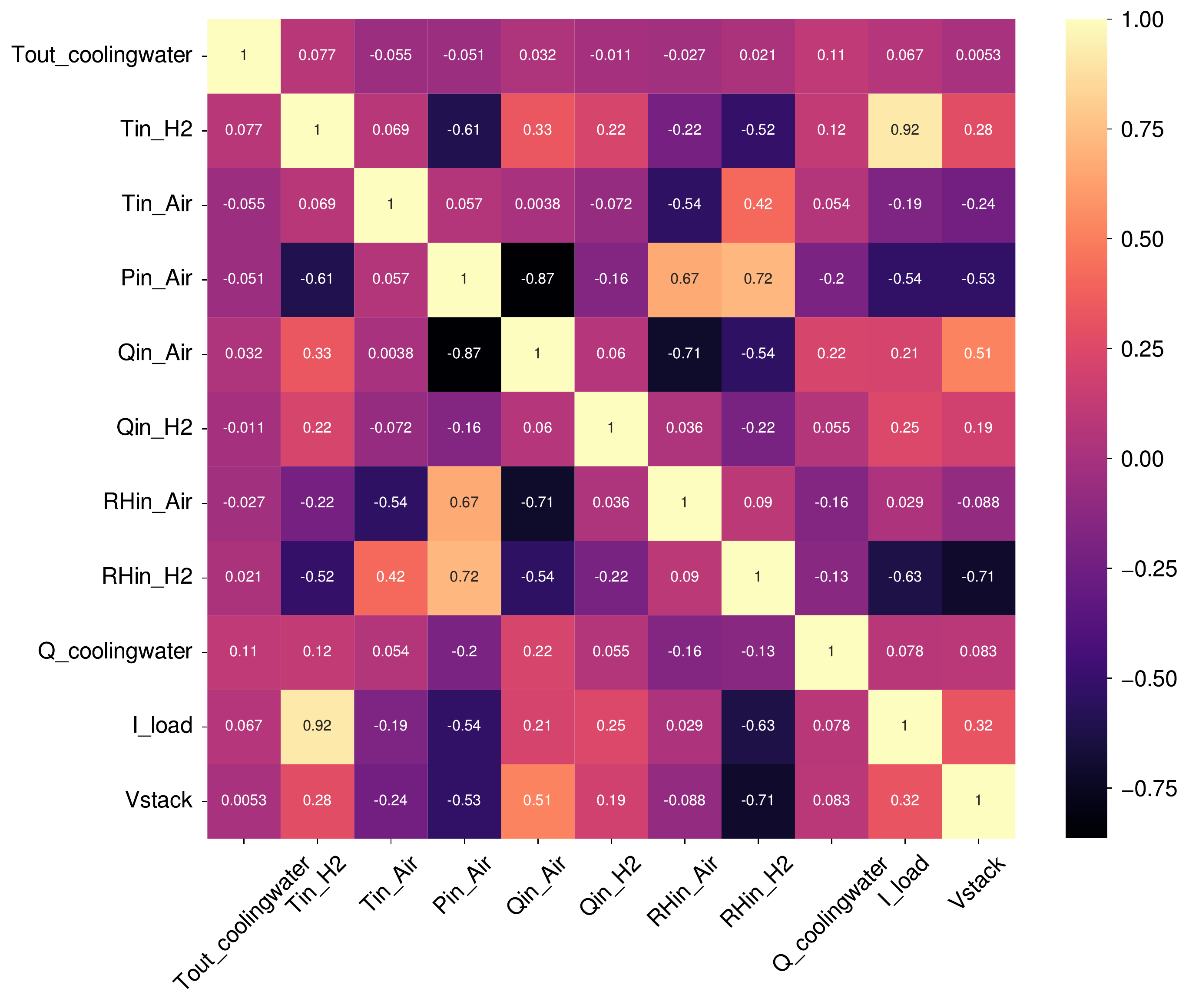} }}
    \caption{Correlation matrices of continuous features (stabilization step).}
    \label{fig:corr_mat}
\end{figure}

The accordance of correlation matrices is the first test to pass for the evaluation of tabular data. The signs of these correlations are corroborated by physical relationships between the variables. Failure by the GAN to reproduce correlations of the same sign would signal failure to learn the structural relations linking the data. The visual representation of those matrices (Figure~\ref{fig:corr_mat}) indicates the absence of any sign error. All the variables are considered in the calculation of the correlations, but for visualization purpose, only a subset are shown: the ten input variables set by the experimental design  along with the stack output voltage (V$_{stack}$). The Kendall similarity score is used to test the similarities in the ordering of data when it is ranked by quantities~\cite{Puka:2011}. It originally lies within $[-1, 1]$ and is mapped in $[0, 1]$, the higher the better. In this first experiment, 100,000 data are generated and the Kendall similarity score is computed for the two matrices. The score is 0.95 (the p-value being less than $1\%$). For all the pairs of variables expected to show a significant correlation (e.g greater than 0.50 in absolute value), no sign error is found and the absolute difference is always less than 0.20.

\subsection{Proximity study}
\label{subsec:prox_study}
\begin{figure}[htp!]
    \centering
    \subfloat[\centering Distributions of critic scores assigned to real and generated data.]{{\includegraphics[width=0.5\textwidth]{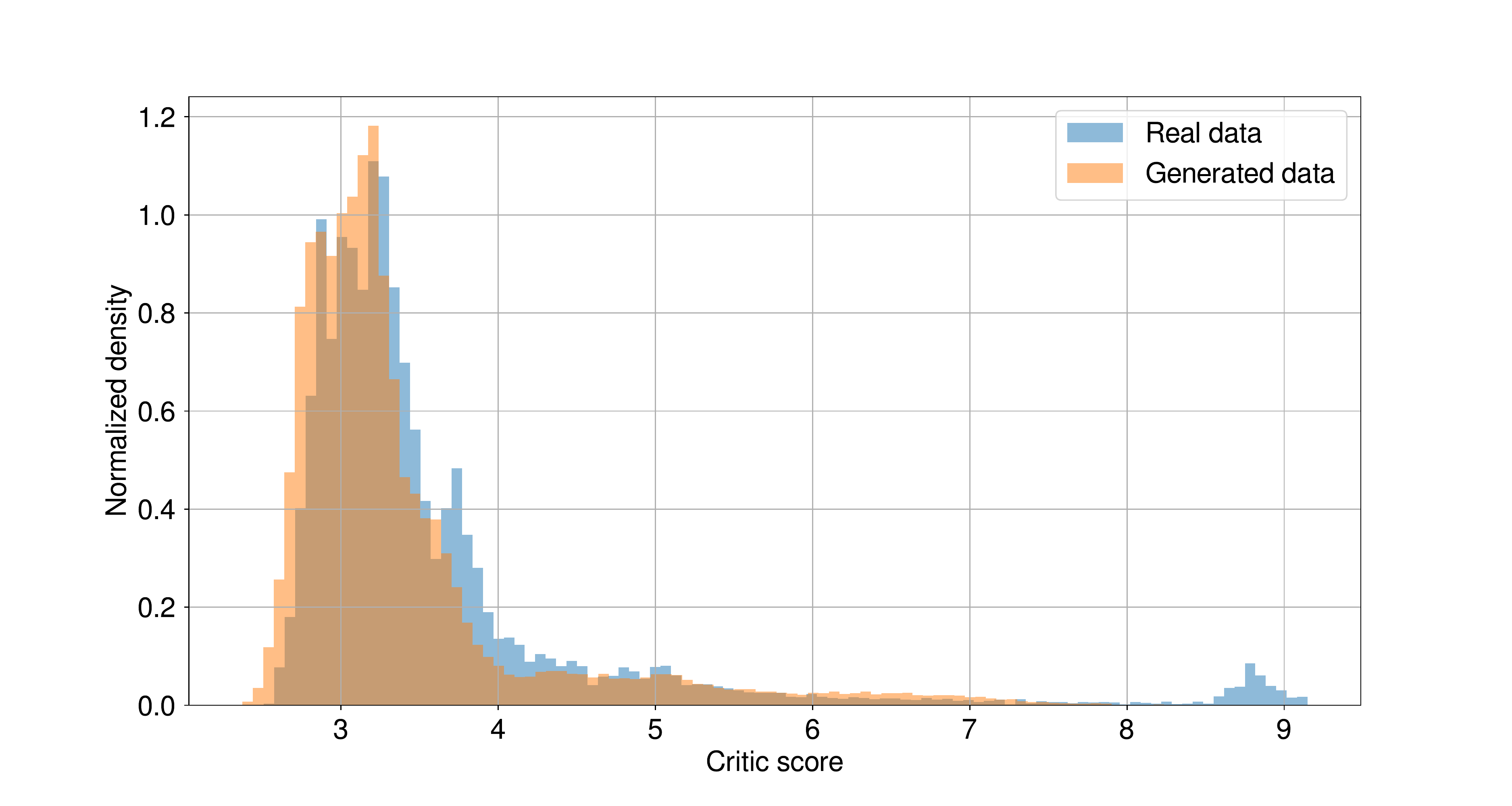}}}
    \subfloat[\centering Proximity score distribution. $\approx 98\%$ of values are kept after removing several outliers. This operation leads to a cropped version of the original distribution within $(-5, 5)$, having a mean of $\mu=-0.20$ and a standard deviation of $\sigma=1.02$).]{{\includegraphics[width=0.5\textwidth]{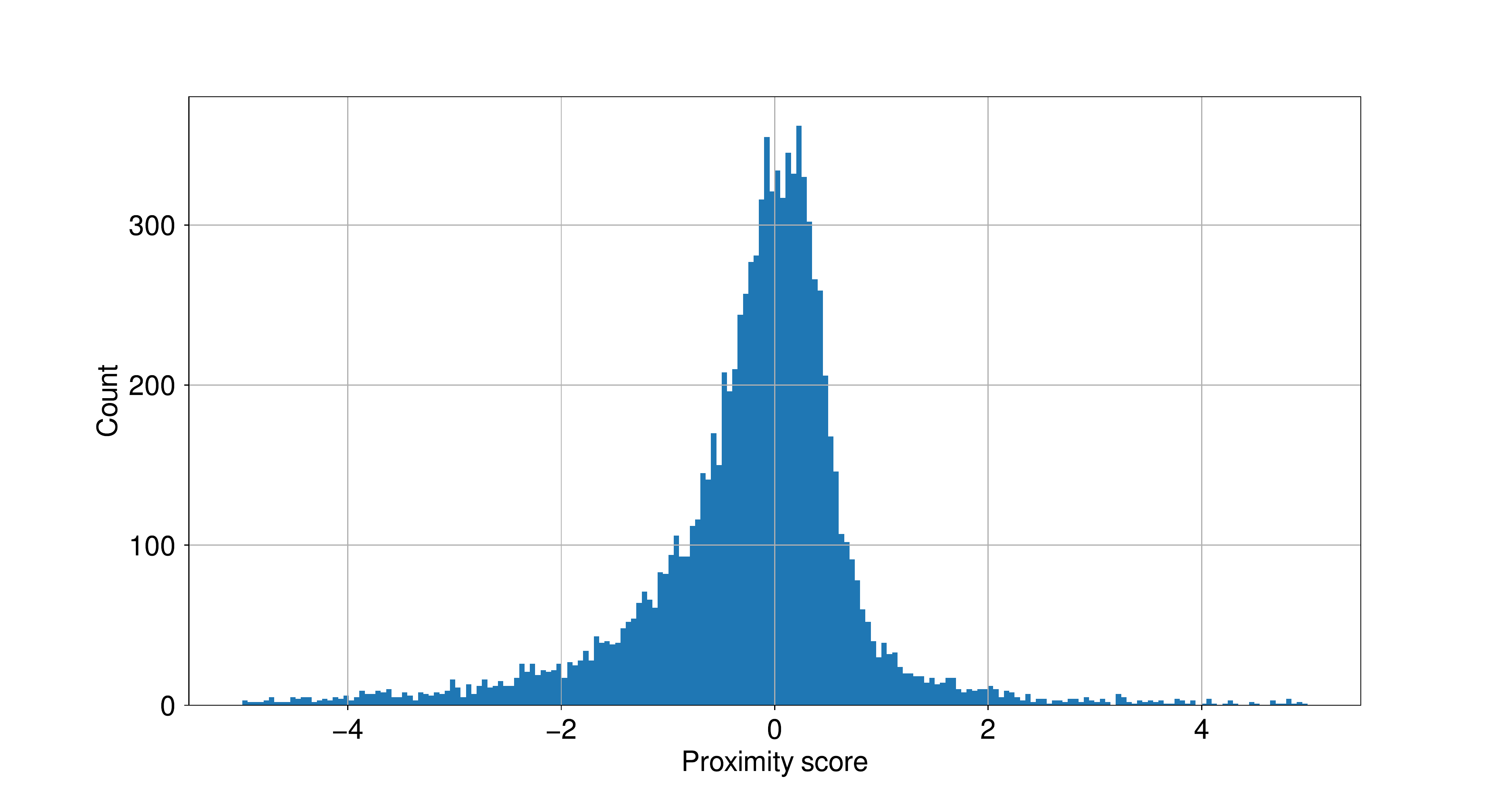}}}
    \caption{Proximity study results (stabilization and polarization steps).}
    \label{fig:proximity_study}
\end{figure}

Another way to appreciate the closeness of the generated data to the real data is to use the critic scores assigned to them. The left part of Figure~\ref{fig:proximity_study} depicts the overlay of the distributions of critic scores assigned to real and 100,000 generated data. Then, a random subset of 10,000 generated data is considered. For each of those vectors of artificial data, the $k=20$ closest real data are found in the training set using the L2 norm restricted to the set of experimental input variables mentioned in the previous section. Then, the GAN critic is called to score the artificial data row and its nearest neighbors selected from the real data. A statistic on the difference over the generated dataset defines a proximity score (the closer to 0 the better). Proximity score distribution is depicted in the right part of Figure~\ref{fig:proximity_study}.

\subsection{Triangle plot}
\label{subsec:tr_plot}
\begin{figure}[htp!]
    \centering
    \includegraphics[width=\linewidth]{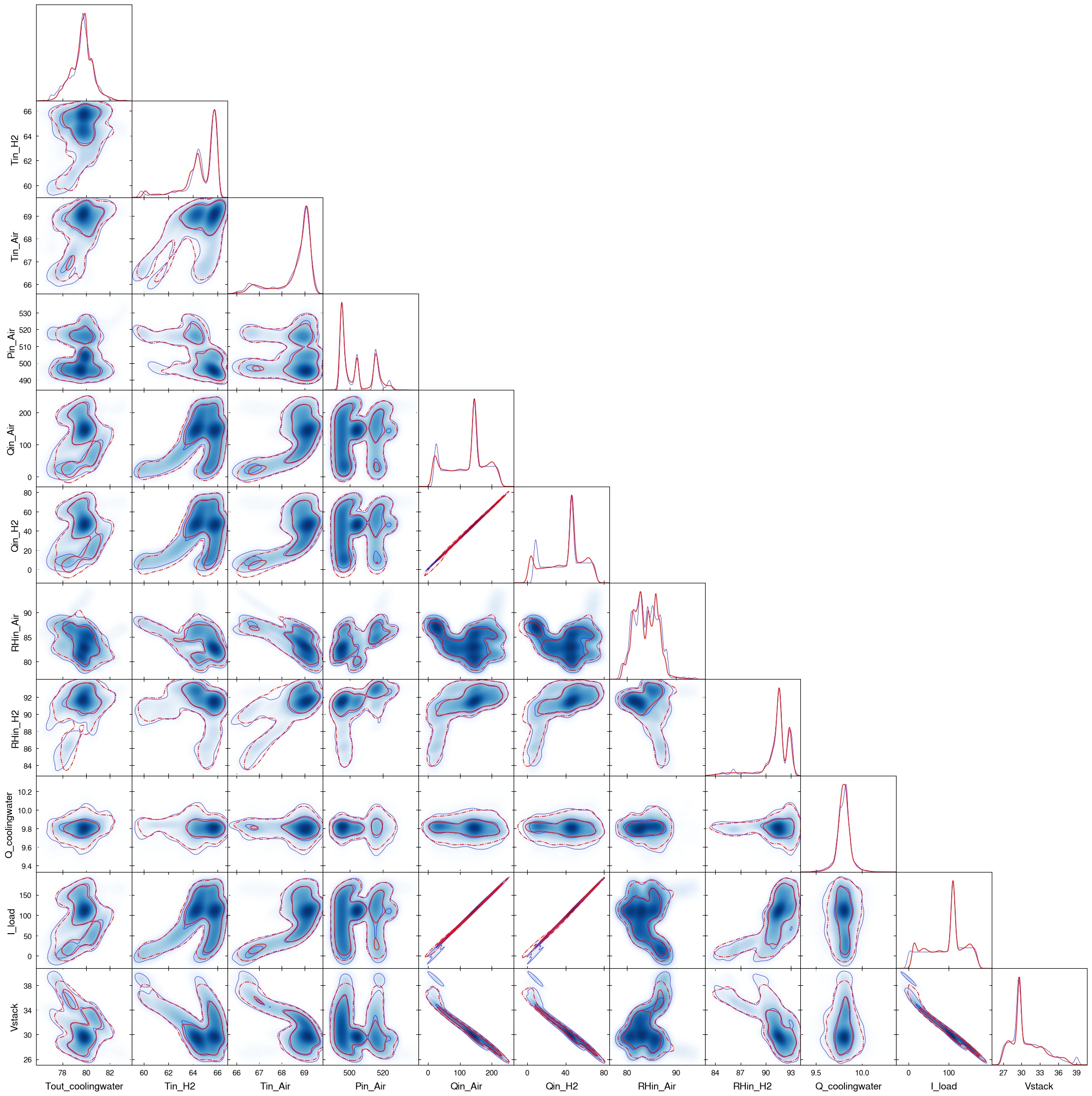}
    \caption{Triangle plot (stabilization and polarization steps): comparison between the real (\textit{blue}) and generated (\textit{red}) datasets via visual inspection of the respective marginal distributions. The one-dimensional (\textit{diagonal panels}) and the two-dimensional (\textit{off-diagonal panels}) marginal distributions are sorted by couples of continuous features. In each subplot, the $68\%$ (\textit{solid line}) and the $95\%$ (\textit{dash-dotted line}) confidence levels are also shown.}
    \label{fig:tr_plot}
\end{figure}

Unlike images, tabular data have no natural visual representation. The graphical representation of the bivariate distributions calculated on all pairs of variables nevertheless provides a visual tool for assessing the closeness between real and generated data. The triangle plot depicted in Figure~\ref{fig:tr_plot} illustrates the replication of univariate modes and complex correlations between variables, and visually extends the comparison of correlation matrices. One can also appreciate the great similarity between the $68\%$ and the $95\%$ confidence levels as an indication of the good performance of the generative model.

\subsection{Polarization curve}
\label{subsec:pola_curve}
\begin{figure}[htp!]
    \centering
    \includegraphics[width=0.95\linewidth]{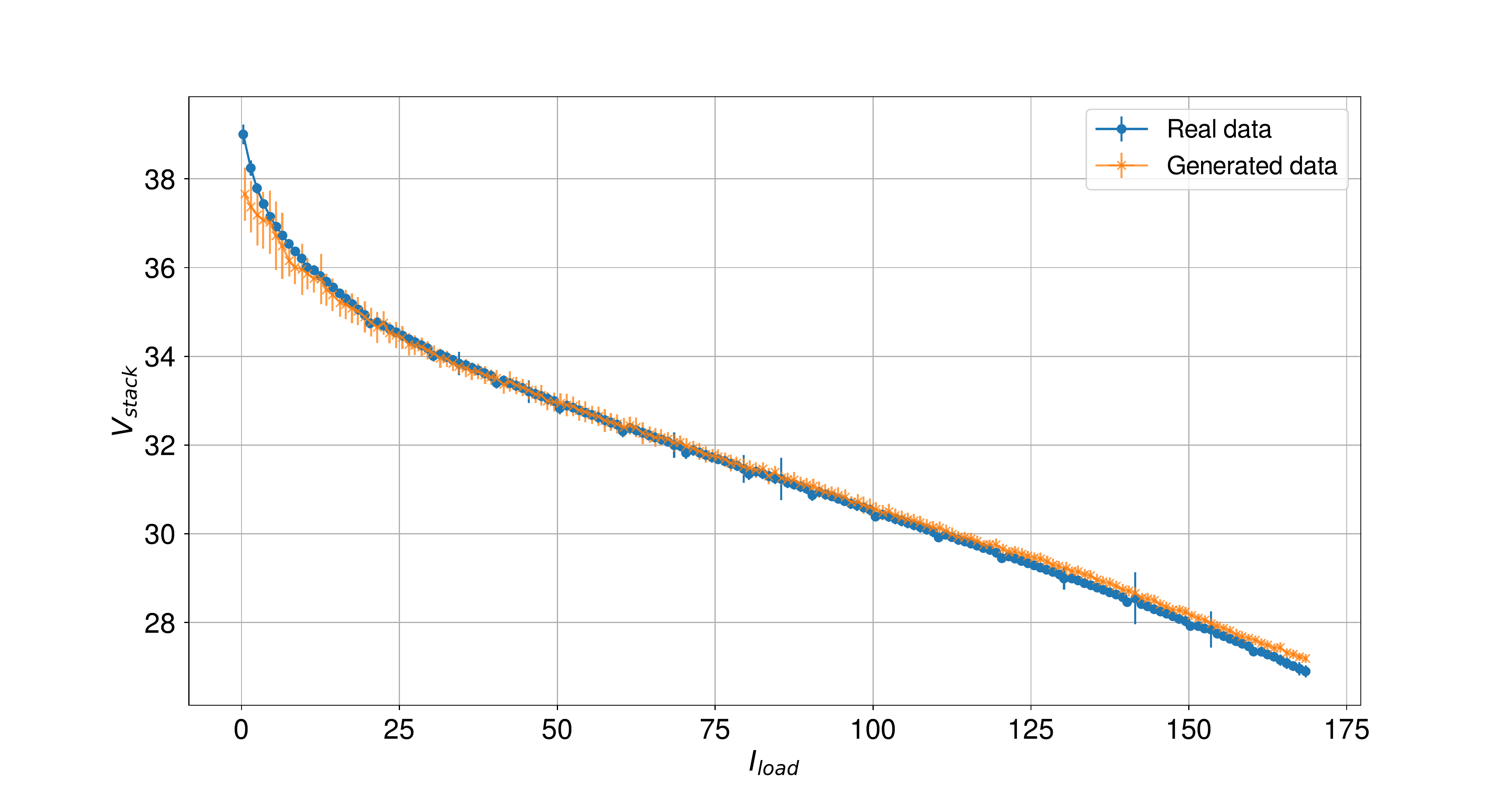}
    \caption{Polarization curve replication ($\approx 25,000$ polarization data vs. $\approx 25,000$ generated data). Polarization closeness errors are equal to $0.41\%$ and $0.66\%$, for $V_{\textrm{stack}}$ and $I_{\textrm{load}}$, respectively. These error rates are to be compared with the measurement errors for cell voltages and load currents which are less than $1\%$. The highest discrepancies appear at both ends of the range, but the non-linear relationship is overall well captured.}
    \label{fig:pola_curve}
\end{figure}

\begin{figure}[htp!]
    \centering
    \includegraphics[width=0.95\linewidth]{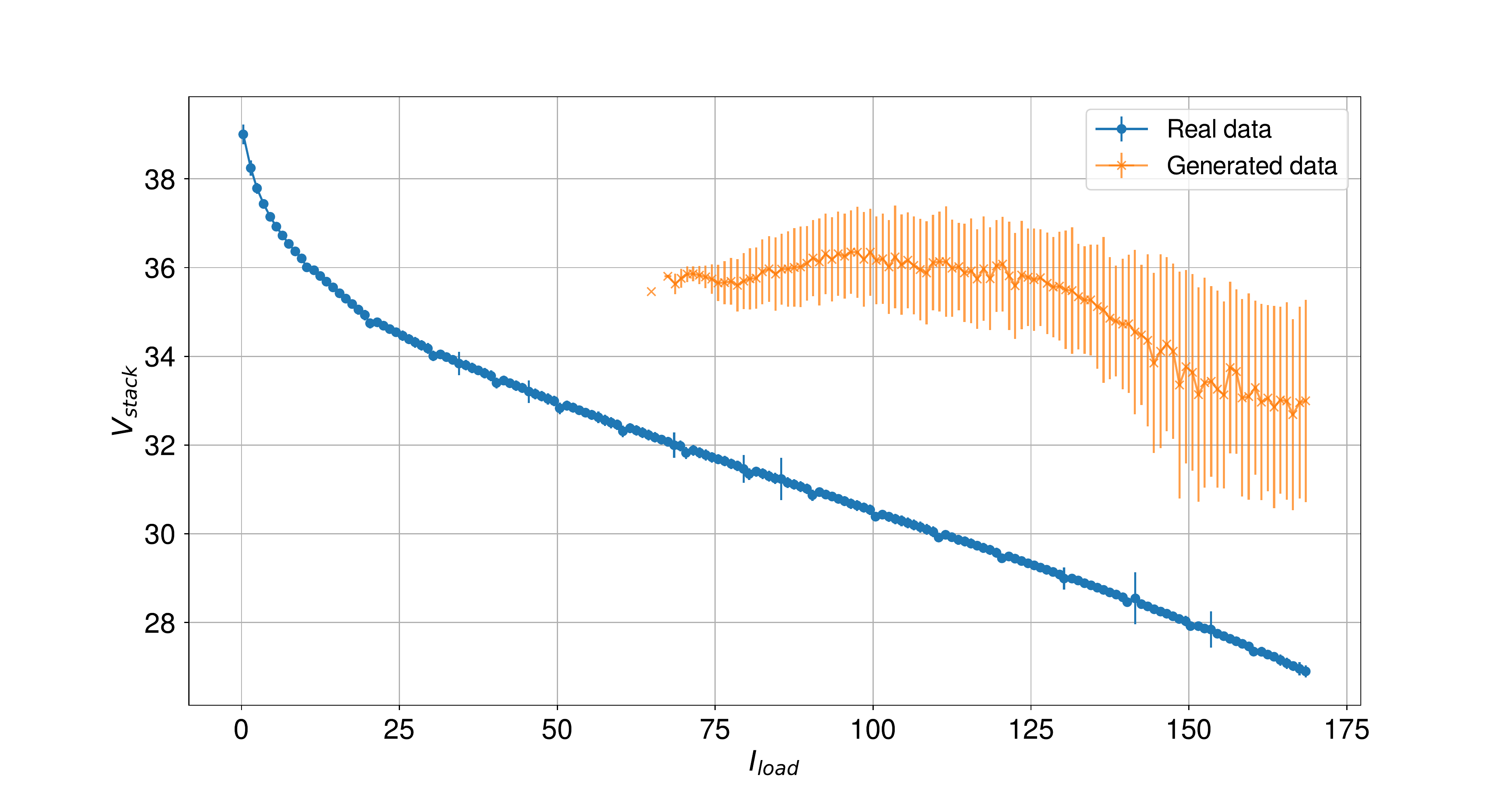}
    \caption{Polarization curve replication ($\approx 25,000$ polarization data vs. $\approx 25,000$ generated data). The same model has been trained only on data points having their load current variable in [110A, 120A]. This is an example of a bad sampling strategy, leading to a polarization closeness error of $18.8\%$ for $V_{\textrm{stack}}$, way beyond the acceptable measurement error for cell voltages of $1\%$.}
    \label{fig:pola_curve_ugly}
\end{figure}

Among the characteristics that we wish to replicate, the most important are the polarization curves. Indeed, these are used routinely to test and troubleshoot fuel cell systems. In order to assess the quality of the reproduction of the polarization curve, a polarization \textit{closeness error metric} is defined (see Section \textit{Methods}). Figures~\ref{fig:pola_curve} and \ref{fig:pola_curve_ugly} illustrate respectively a very good and poor replication of the polarization curve from the GAN. In Figure~\ref{fig:pola_curve_ugly}, the learning set has been filtered so as to only keep observations having the variable $I_{load}$ within $[110A, 120A]$.
One important finding of this experiment is about data distribution: the GAN can be successfully trained on a reduced set of data, as shown in the subsampling study (Table~\ref{tab:sens_study}), as long as the training data are sufficiently well distributed in space. The GAN shows in this case an interesting capacity to interpolate between experimental points. Conversely, when the training is restricted to a small region of the space, the GAN fails to reproduce the data, even within the training area.

\subsection{Subsampling study}
\label{subsec:subsamp_study}
The objective of this study is to analyze the impact of the number of real data available on the performance of the GAN. The real data are randomly and uniformly selected from the original dataset (preserving the original distribution for both classes -- $20\%$ for stabilization, $80\%$ for polarization) and serve for training the GAN. This operation is equivalent to gradually lowering the original data acquisition factor, here from 1 down to 1/4. Performance metrics are shown in Table~\ref{tab:sens_study}. What we can learn from this study is that, despite some statistical fluctuations, those performance metrics remain high, proving that the GAN does not need as much data as the original dataset (five days of experiments) to generate artificial data with a high level of similarity.

\begin{table}[htp!]
\centering
\begin{tabular}{@{}lccc@{}}
\toprule
Data acquisition factor             & $f=1$        & $f=1/2$      & $f=1/4$ \\
\midrule
Real data (1)                       & 30,901       & 15,450       & 7,725 \\
Generated data (2)                  & 100,000      & 100,000      & 100,000 \\
Ratio (2) / (1)                     & 3.24$\times$ & 6.47$\times$ & 12.9$\times$ \\
Epochs                              & 25,000       & 25,000       & 25,000 \\
Batch size                          & 256          & 256          & 256 \\
Training time (min)                 & 60.2         & 60.0         & 60.8 \\
KS score                            & 0.87 ($\pm 1.5\mathrm{e}{-3}$) & 0.88 ($\pm 1.5\mathrm{e}{-3}$) & 0.87 ($\pm 1.0\mathrm{e}{-3}$) \\
Dim red score                       & 0.97 ($\pm 4.1\mathrm{e}{-3}$) & 0.91 ($\pm 6.4\mathrm{e}{-3}$) & 0.97 ($\pm 1.7\mathrm{e}{-5}$) \\
Kendall score                       & 0.90 ($\pm 3.5\mathrm{e}{-4}$) & 0.89 ($\pm 1.9\mathrm{e}{-4}$) & 0.90 ($\pm 2.3\mathrm{e}{-4}$) \\
Pola error (\%)                     & 0.64 ($\pm 2.6\mathrm{e}{-5}$) & 0.44 ($\pm 2.0\mathrm{e}{-5}$) & 0.65 ($\pm 2.8\mathrm{e}{-5}$) \\
\bottomrule
\end{tabular}
\caption{Performance results for the subsampling study on stabilization and polarization steps. In each experiment, one GAN model is trained and 100,000 data are generated and compared to the original dataset (with 30,901 real data). Performance metrics are given in terms of means and standard deviations over 100 different inferences. The Kendall score is mapped in the range [0, 1] and the polarization closeness error metric (see Section \textit{Methods}) utilizes polarization data only.}
\label{tab:sens_study}
\end{table}

\section{Discussion}
\label{subsec:discussion}
In fuel cell systems, many parameters must be set in specific ranges in order to achieve an optimal performance of the device, such as the reactants' temperatures, pressures, flows, and hydration levels, the stack temperature, the current, and the voltage. One of the challenges of this study is to find relationships between the artificial variables that are dictated by the underlying physics. For instance, the gas flow is proportional to the current.\\
The proposed GAN architecture produces artificial data that faithfully reproduces the joint distribution of the real ones. A high replication quality is achieved, as illustrated by the calculation of several statistical metrics. It is however important to underline the strong dependence of the resulting generator on the experimental design used to produce the data. The analysis of the joint distributions shows that for each pair of input variables, only a small fraction of the respective ranges of variation was explored.  There are at least two reasons for this. The first is related to the number of acquisitions. If for every input variable their entire range of values must be explored, the total amount of acquisitions would increase exponentially which is clearly unfeasible. The second reason is physical and involves the experimenter's know-how: not all combinations of variables are valid experimental conditions, some of them not allowing the stabilization of the stack, or even leading to physical degradations. The experimental design appears therefore far from arbitrary. If it is improperly chosen, like in the example presented for the Polarization Curve, the GAN may lose its ability to replicate data. The main strength of the current generator is its ability to interpolate data within the multi-dimensional experimental space, thus reducing the time spent by the fuel cells on the test bench.\\
As mentioned above, the bivariate plots show a visually very high degree of replication of the experimental design by the GAN (Figure~\ref{fig:tr_plot}). For example, the marginal distribution of the couple of variables ($V_{\textrm{stack}}$, $I_{\textrm{load}}$) is captured with a very good precision. In order to be generalized in higher dimension, this visual fit can be translated into image similarity scores. In this particular case, the most important physical relationships involve at most two variables, which makes the verification of the trivariate marginal distributions less crucial, keeping in mind that this verification has cubic complexity with respect to the number of variables. \\
The polarization curve is replicated with a relative error that is, on average, less than the measurement uncertainty (Figure~\ref{fig:pola_curve}). A maximum distortion of 1.5\% appears on the output voltage in the region of low currents (less than 5A). This distortion is very likely to be 
reduced by further improvements in the GAN architecture.\\
Considering the objective of reducing as much as possible the time spent on the test bench and the associated costs, it is possible to go from 4 hours of measurements (relative to one polarization curve) to almost one hour. It is important to note that once the model is trained, it only takes less than 35 ms to generate a polarization curve with 5,000 points . This guarantees a considerable time saving when evaluating the performances of a fuel cell stack.

\section{Conclusion}
\label{sec:ccl}
The fast generation of reliable and consistent artificial data paves the way to promising developments and operational improvements in fuel cell qualification. According to Table \ref{tab:sens_study}, the characterization time can be conservatively cut by 4. A next step will be the extension of the generator to non-nominal operating modes (e.g. air and fuel varying stoichiometries, high and low operating temperatures) to ensure the ability to map the behavior of the fuel cell stack in all operating conditions. The expected benefits of data generation are even greater in default mode than in normal operation, since experiments under fault conditions degrade the equipment and can therefore only be performed a limited number of times. A further step will be to use these artificial data collected in normal and default modes to train a model of the fuel cell performance throughout its lifetime. 

\section{Methods}
\label{sec:methods}

\subsection{Data from experimental tests}
\label{subsec:exp_data}

We consider a proton exchange membrane fuel cell (PEMFC) with an active area of 220cm$^2$, a nominal operating temperature of $80^\circ C$, and a nominal current density of 0.5A/cm$^2$. It is composed of 40 cells and delivers up to 3kW of electric power. On the test bench, various parameters are supervised and regulated such as the pressures, the flow rates, the gas humidity, and the dew points. The data generation is based on several experimental tests done on the fuel cell stack, in various operating conditions. First, an experiment under nominal operating conditions is performed. To evaluate the performances of the fuel cell stack, a polarization curve (voltage vs. current curve) is performed. Then, it is required to investigate the behavior of the fuel cell stack under degraded operating conditions. Thus, the values of the parameters are progressively increased or decreased up to a given threshold, defined by the physical limits of the test bench and by the fuel cell specifications. 
The full generated dataset is thus divided into several parts: $i)$ start-up of the fuel cell, $ii)$ stabilization of the fuel cell at nominal operating conditions, $iii)$ polarization curves and $iv)$ shutdown. The start-up time of the fuel cell depends on its subsequent use. Indeed, if the fuel cell is brand new, it will be necessary to break it in. Even if some optimization methods can be applied to the fuel cell manufacturing process to reduce this time, it will take multiple hours. If the time to perform a polarization curve is added to the "break-in" process then this will further increase the time on the test bench. To consider a mass production of fuel cells at low cost, it is mandatory to reduce the time spent per fuel cell on a test bench or to use the concept presented in this paper. After data merging, this currently represents approximately 4 hours.

\subsection{GANs architecture}
\label{subsec:gans_archi}
Generative Adversarial Networks (GANs) were first introduced\cite{Goodfellow:2014} in 2014 by Ian Goodfellow. They belong to the family of generative models and are still considered to be the state of the art for approximating real data distributions. They allow to sample multivariate probability distributions with implicit density. GANs involve two actors: one is called the generator, the other the discriminator. The goal of the generator is to create fake samples that are supposed to come from the same distribution as the original training data. The discriminator tries to determine whether its input data are real or fake. The competition in this game pushes both entities to improve themselves until the fake data are indistinguishable from the real ones. This sweet spot is called the Nash equilibrium\cite{Nash:1950} and corresponds to the moment when the model has completed its training. In practice, the generator and the discriminator are represented by differentiable functions such as a deep neural networks. In this work, we adopt as a blueprint a Wasserstein-GAN with gradient penalty (WGAN-GP)\cite{Gulrajani:2017} (the discriminator is call the critic) in which we customize the architecture to handle continuous and categorical variables simultaneously. The WGAN-GP brings strong modeling performance and more stability for training compared to its predecessors. Moreover, it has a loss function based on the concept of optimal transport \cite{Villani:2003, Kolouri:2017}, well suited to approximate the shape of statistical distributions which correlates with generated samples quality. Detailed structure of both the generator and the critic are available in Table~\ref{tab:gan_archi_gen} and Table~\ref{tab:gan_archi_crit} from the appendix, respectively. Regarding the generator architecture, The sub-module dedicated to the processing of a categorical variable is kept generic, indicating with $k_i$ the number of classes for the $i^{\mathrm{th}}$ categorical variable. At training time, the output layer of the generator is a concatenation of the layers B and C$_{i=1,\dots,N}$, for a total dimension of $D=d+\sum_{i=1}^{N}k_i$, $d$ being the number of non-categorical variables. The abbreviation \textit{Conn.} stands for \textit{Connected}. As far as the critic architecture is concerned, the abbreviation \textit{dr.} stands for \textit{dropout rate}.

\subsection{GANs training parameters}
\label{subsec:gans_train_params}
The model is trained for 25,000 epochs. It was verified that without adding any further complexity to the overall architecture, pushing the training for more epochs was useless and could even lead to worse performance results. A batch size of $n_{\mathrm{b}}=256$ data is chosen as a good compromise between rapidity and quality of the training. Adam optimizer \cite{Kingma:2017} is used with a learning rate of $l_{\mathrm{r}}=0.001$ for both the generator and the critic. The dimension of the latent space is set to $L_{\mathrm{dim}}=150$ which, even if it is higher than the number of variables, helps to improve performance results by decreasing the distortion of the generated data, as mentioned in \cite{Pinetz:2020}. The gradient penalty is set to $\lambda=10$, in line with previous literature \cite{Kingma:2017} while the critic / generator training ratio is equal to 15. The total number of parameters of the model is 218,418 (120,769 for the generator and 97,649 for the critic).

\subsection{Kolmogorov-Smirnov score}
\label{subsec:ks_score}
The two-sample Kolmogorov–Smirnov (KS) test \cite{Hodges:1958} is used to verify whether two one-dimensional samples come from the same distribution. This metric compares the distributions of continuous features using the empirical cumulative distribution function (CDF). For each feature, the similarity score is computed as one minus the KS test D-statistic, which indicates the maximum distance between the real data CDF ($CDF_{\mathrm{real}}$) and the generated data CDF ($CDF_{\mathrm{gen}}$) values. The output score $S_{\mathrm{ks}}$ lies in $[0, 1]$ (the higher the better) and is computed as the minimum score across all the features:
\begin{equation}
    S_{\mathrm{ks}} = \min_{i} (1-\sup_{x_i}|CDF_{\mathrm{real}}(x_i)-CDF_{\mathrm{gen}}(x_i)|)
\end{equation}
The minimum is taken instead of the mean that would smooth errors. Thus, if this score is low, we would have the information that at least one feature has not been approximated correctly. This metric quantitatively complements the visual inspection of the triangle plot (Figure~\ref{fig:tr_plot}) to assess the quality of the reproduction of marginal distributions.

\subsection{Dimension reduction score}
\label{subsec:dim_red_score}
The dimension reduction score aims to reflect how well the model has captured the correlations between the different features. First, a Principal Component Analysis (PCA) \cite{Pearson:1901} is performed on both real and generated data such that $99\%$ of the variance is explained by the first $n$ eigenvectors. Then, real and generated eigenvectors are normalized by the square root of their corresponding eigenvalues (resp. $m_{\mathrm{real}}$ and $m_{\mathrm{gen}}$) before a cosine similarity $S_{\mathrm{cosine}}$ is computed between them. The output score $S_{\mathrm{dim}}$ lies within $[0, 1]$ (the higher the better) and is computed as follows:
\begin{equation}
    S_{\mathrm{dim}}=\frac{\displaystyle \sum_{i=1}^{n}{\omega_{i} \times S_{\mathrm{cosine}}^{i}\left(m_{\mathrm{real}}^{i}, m_{\mathrm{gen}}^{i}\right)}}{\displaystyle \sum_{i=1}^{n}{\omega_{i}}}, i=1 \ldots n
\label{eq:dim_red_score}
\end{equation}
with
\begin{equation}
    S_{\mathrm{cosine}}(x, y)=\frac{1}{2}\left( \frac{x \cdot y}{\lvert x \rvert \times \lvert y \rvert}+1 \right) \in [0, 1]
\label{eq:cos_score}
\end{equation}
where $\omega_{i}$ corresponds to the explained variance ratio of the $i$th real eigenvector.

\subsection{Proximity score}
\label{subsec:prox_score}
In the neighborhood study, the critic score from one generated row $\mathrm{S_{c}^{gen}}$ is compared to the distribution of the critic scores of its $k$-nearest neighbors $\mathrm{S_{c}^{knn}}$ using a proximity score $\mathrm{S_{prox}^{g}}$ computed as follows:

\begin{equation}
    \mathrm{S_{prox}^{g}} = \frac{\mathrm{S_{c}^{gen}} - \min{\left\{\mathrm{S_{c}^{knn}}\right\}}}{\max{\left\{\mathrm{S_{c}^{knn}}\right\}} - \min{\left\{\mathrm{S_{c}^{knn}}\right\}}}
    \label{eq:proximity_score}
\end{equation}

\subsection{Polarization closeness error}
\label{subsec:pola_error}
To compute the polarization closeness error, we divide the currents in $nc=\max{( \lceil I_{load} \rceil )}$ bins, and compute the average of the currents and of the voltages for both real and generated data in each bin (which we will note $\overline{V}_{\mathrm{real}, i}$, $\overline{V}_{\mathrm{gen}, i}$, $\overline{I}_{\mathrm{real}, i}$, and $\overline{I}_{\mathrm{gen}, i}$). This allows us to produce the box plots visible in Figure~\ref{fig:pola_curve}. We then take the average of the relative error between real and generated data in each bin, which gives us the average percentage error $e$ between both data types :
\begin{equation}
    e = \frac{1}{nc} \sum_{i = 1}^{nc} \bigg | \frac{\overline{V}_{\mathrm{real}, i} - \overline{V}_{\mathrm{gen}, i}}{\overline{V}_{\mathrm{real}, i}} \bigg |
\end{equation}
\section*{Acknowledgements}

This work has been supported by the EIPHI Graduate School (contract ANR-17-EURE-0002) and the Region Bourgogne Franche-Comté
\appendix

\section{Experimental Data Information}
\label{sec:appendix_exp_data_info}
\begin{landscape}
\begin{table}[h]
\footnotesize
\centering
\begin{tabular}{|l|l|l|l|c|}
\hline
Feature name & Description & Unit & Range & Histogram\\
\hline
Tout$_{cooling water}$ & Outlet temperature of the water & $^{\circ}$C & [25 - 95] & \includegraphics[width=0.2\textwidth, height=10mm]{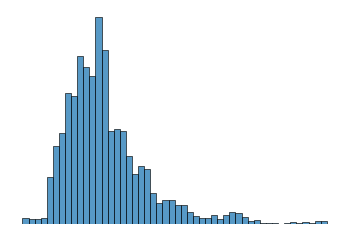}\\
\hline
Tin$_{H_2}$ & Inlet temperature of the hydrogen & $^{\circ}$C & [25 - 95] & \includegraphics[width=0.2\textwidth, height=10mm]{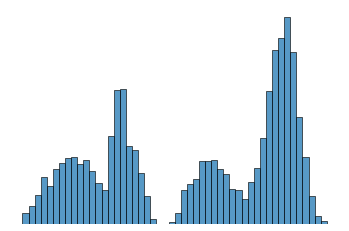}\\
\hline
Tin$_{Air}$ & Inlet temperature of the air & $^{\circ}$C & [25 - 95] & \includegraphics[width=0.2\textwidth, height=10mm]{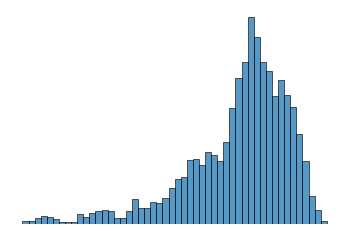}\\
\hline
Pin$_{Air}$ & Inlet pressure of the air & mbarg & [0 - 1000] & \includegraphics[width=0.2\textwidth, height=10mm]{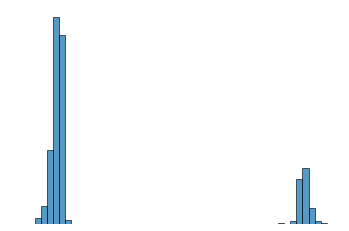}\\
\hline
Pin$_{H2}$ & Inlet pressure of the hydrogen & mbarg & [0 - 1000] & \includegraphics[width=0.2\textwidth, height=10mm]{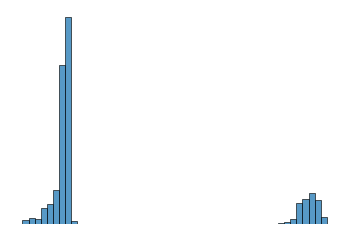}\\
\hline
Qin$_{Air}$ & Total inlet air flow rate & Nl/min & [0 - 300] & \includegraphics[width=0.2\textwidth, height=10mm]{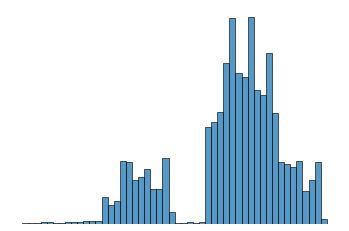}\\
\hline
Qin$_{H_2}$ & Total inlet hydrogen flow rate & Nl/min & [0 - 100] & \includegraphics[width=0.2\textwidth, height=10mm]{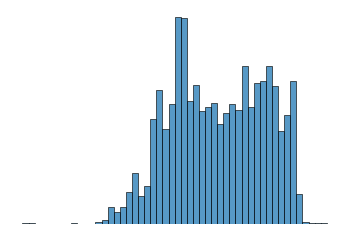}\\
\hline
RHin$_{Air}$ & Relative humidity of the air (inlet of the stack) & \% & [0 - 100] & \includegraphics[width=0.2\textwidth, height=10mm]{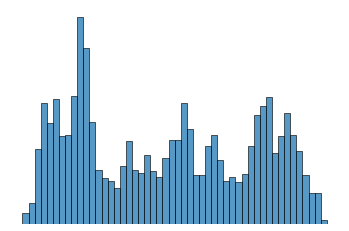}\\
\hline
RHin$_{H2}$ & Relative humidity of the hydrogen (inlet of the stack) & \% & [0 - 100] & \includegraphics[width=0.2\textwidth, height=10mm]{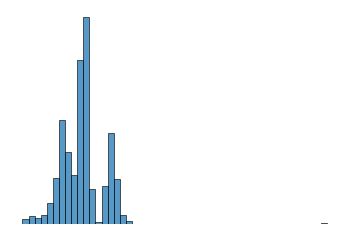}\\
\hline
Q$_{cooling water}$ & Deionized water flow rate & l/min & [0 - 15] & \includegraphics[width=0.2\textwidth, height=10mm]{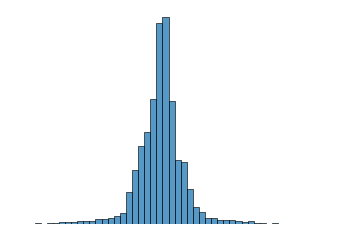}\\
\hline
V$_{stack}$ & Stack voltage & V & [0 - 40] & \includegraphics[width=0.2\textwidth, height=10mm]{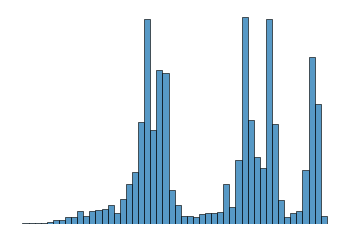}\\
\hline
I$_{load}$ & Current load & A & [0 - 200] & \includegraphics[width=0.2\textwidth, height=10mm]{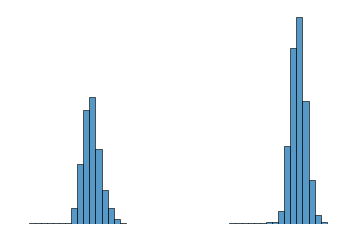}\\
\hline
Vcell$_{i}$ & Cell voltage of the cell $\#i$ ($i=1\ldots40$) & V & [0 - 1] & \includegraphics[width=0.2\textwidth, height=10mm]{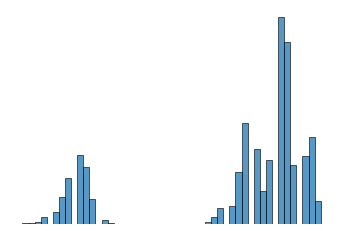}\\
\hline
\end{tabular}
\caption{\label{tab:data}Description of the continuous real data involved in the PEMFC experiments (stabilization step).}
\end{table}
\end{landscape}

\section{GANs Architecture}
\label{sec:appendix_gans_archi}

\setlength{\tabcolsep}{2pt}
\begin{table}[htp!]
    \footnotesize
    \centering
    \setlength{\doublerulesep}{2\arrayrulewidth}
    \begin{tabular}{l c c r}
    \toprule
    \cmidrule(lr){1-4}
    Layer type  & Output shape & Parameters  & Info.\\
    \cmidrule{1-4}\morecmidrules\cmidrule{1-4}
    \multicolumn{4}{c}{Input sub-module}\\
    \cmidrule(lr){1-4}
    Input &  ($n_{\mathrm{b}}$, 150)  & 0   &  \\
    \cdashlinelr{1-4}
    Dense & ($n_{\mathrm{b}}$, 256)  & 38656  & \\
    Batch Norm.  & ($n_{\mathrm{b}}$, 256)  & 1024  & \\
    $(s=+1)$-SLR  &($n_{\mathrm{b}}$, 256) & 2   & \\
    \cdashlinelr{1-4}
    Dense & ($n_{\mathrm{b}}$, 128)  & 32896  & \\
    Batch Norm.  & ($n_{\mathrm{b}}$, 128)  & 512  & \\
    $(s=-1)$-SLR   &($n_{\mathrm{b}}$, 128) & 2  & \\
    \cdashlinelr{1-4}
    Dense &   ($n_{\mathrm{b}}$, 64)  & 8256  & \\
    Batch Norm.  & ($n_{\mathrm{b}}$, 64)  & 256  & \\
    $(s=+1)$-SLR    &($n_{\mathrm{b}}$, 64) & 2 & \\
    \cdashlinelr{1-4}
    Dense &   ($n_{\mathrm{b}}$, 32)  & 2080  & \\
    Batch Norm.  & ($n_{\mathrm{b}}$, 32)  & 128  & \\
    $(s=-1)$-SLR    &($n_{\mathrm{b}}$, 32) & 2 & Layer A\\
    \cmidrule(lr){1-4}
    \multicolumn{4}{c}{$d$ continuous variables}\\
    \cmidrule(lr){1-4}
    Dense &  ($n_{\mathrm{b}}$, $4d$)  &  $4d\cdot 33$ & \shortstack{Conn. to A}\\
    Batch Norm.  & ($n_{\mathrm{b}}$, $4d$)  & $4d\cdot 4$ & \\
    $(s=-1)$-SLR    &($n_{\mathrm{b}}$, $4d$) & 2  & \\
    \cdashlinelr{1-4}
    Dense &  ($n_{\mathrm{b}}$, $2d$)  & $(4d+1)\cdot 2d$ & \\
    Batch Norm.  &($n_{\mathrm{b}}$, $2d$)  & $2d\cdot 4$ &  \\
    $(s=+1)$-SLR    &($n_{\mathrm{b}}$, $2d$) & 2 &  \\
    \cdashlinelr{1-4}
    Dense &  ($n_{\mathrm{b}}$, $d$)  & $(2d+1)\cdot d$ & Layer B\\
    \cmidrule(lr){1-4}
    \multicolumn{4}{c}{ $i^{\mathrm{th}}$ categorical variable of $k_i$ classes ($i=1,\dots,N$)}\\
    \cmidrule(lr){1-4}
    Dense & ($n_{\mathrm{b}}$, $4k_i$)  & $4k_i\cdot 33$ & \shortstack{Conn. to A}\\
    Batch Norm.  &($n_{\mathrm{b}}$, $4k_i$)  & $4k_i\cdot 4$ & \\
    LeakyReLU & ($n_{\mathrm{b}}$, $4k_i$) & 0  & \\
    \cdashlinelr{1-4}
    Dense & ($n_{\mathrm{b}}$, $2k_i$)  & $(4k_i+1)\cdot2k_i$ & \\
    Batch Norm.  & ($n_{\mathrm{b}}$, $2k_i$)  & $(4k_i+1)\cdot4$ & \\
    LeakyReLU & ($n_{\mathrm{b}}$, $2k_i$) & 0  & \\
    \cdashlinelr{1-4}
    Dense & ($n_{\mathrm{b}}$, $k_i$)  &  $(2k_i+1)\cdot k_i$ & \\
    Batch Norm.  &($n_{\mathrm{b}}$, $k_i$)  & $(2k_i+1)\cdot4$ & \\
    \cdashlinelr{1-4}
    Softmax & ($n_{\mathrm{b}}$, $k_i$)  & 0 & Layer C$_i$\\
    \cmidrule(lr){1-4}
    Concatenate & ($n_{\mathrm{b}}$, $D$) & 0 & Output (B, C$_{i=1, \dots, N})$\\
    \bottomrule
     \end{tabular}
    \caption{Generator architecture as function of the batch size $n_{\mathrm{b}}$.}
    \label{tab:gan_archi_gen}
\end{table}

\setlength{\tabcolsep}{2pt}
\begin{table}[htp!]
    \footnotesize
    \centering
    \setlength{\doublerulesep}{2\arrayrulewidth}
    \centering
    \renewcommand{\arraystretch}{1.046}
    \begin{tabular}{l c c r}
    \toprule
    \cmidrule(lr){1-4}
    Layer type  & Output shape & Parameters  & Info.\\
    \cmidrule{1-4}\morecmidrules\cmidrule{1-4}
    Input & ($n_{\mathrm{b}}$, 59)  & 0  &  \\
    \cdashlinelr{1-4}
    Dense & ($n_{\mathrm{b}}$, 256)  & 15360 & \\
    LeakyReLU &($n_{\mathrm{b}}$, 256)  & 0 & \\
    Dropout & ($n_{\mathrm{b}}$, 256) & 0 & No \\
    \cdashlinelr{1-4}
    Dense & ($n_{\mathrm{b}}$, 128)  & 32896 & \\
    LeakyReLU & ($n_{\mathrm{b}}$, 128)  & 0 & \\
    Dropout & ($n_{\mathrm{b}}$, 128) & 0 & No \\
    \cdashlinelr{1-4}
    Dense & ($n_{\mathrm{b}}$, 128)  & 16512 & \\
    LeakyReLU  &($n_{\mathrm{b}}$, 128)  & 0 & \\
    Dropout & ($n_{\mathrm{b}}$, 128) & 0 & No \\
    \cdashlinelr{1-4}
    Dense & ($n_{\mathrm{b}}$, 128)  & 16512 & \\
    LeakyReLU & ($n_{\mathrm{b}}$, 128)  & 0 & \\
    Dropout & ($n_{\mathrm{b}}$, 128) & 0 & dr=0.5 \\
    \cdashlinelr{1-4}
    Dense & ($n_{\mathrm{b}}$, 64)  & 8256 & \\
    LeakyReLU & ($n_{\mathrm{b}}$, 64)  & 0 & \\
    Dropout & ($n_{\mathrm{b}}$, 64) & 0 & dr=0.5 \\
    \cdashlinelr{1-4}
    Dense & ($n_{\mathrm{b}}$, 64)  & 4160 & \\
    LeakyReLU & ($n_{\mathrm{b}}$, 64)  & 0 & \\
    Dropout & ($n_{\mathrm{b}}$, 64) & 0 & dr=0.2 \\
    \cdashlinelr{1-4}
    Dense & ($n_{\mathrm{b}}$, 32)  & 2080 & \\
    LeakyReLU & ($n_{\mathrm{b}}$, 32)  & 0 & \\
    Dropout &($n_{\mathrm{b}}$, 32) & 0 & dr=0.2 \\
    \cdashlinelr{1-4}
    Dense &($n_{\mathrm{b}}$, 32)  & 1056 & \\
    LeakyReLU  &($n_{\mathrm{b}}$, 32)  & 0 & \\
    Dropout & ($n_{\mathrm{b}}$, 32) & 0 & No \\
    \cdashlinelr{1-4}
    Dense & ($n_{\mathrm{b}}$, 16)  & 528 & \\
    LeakyReLU & ($n_{\mathrm{b}}$, 16)  & 0 & \\
    Dropout & ($n_{\mathrm{b}}$, 16) & 0 & No \\
    \cdashlinelr{1-4}
    Dense & ($n_{\mathrm{b}}$, 16)  & 272 & \\
    LeakyReLU & ($n_{\mathrm{b}}$, 16)  & 0 & \\
    Dropout & ($n_{\mathrm{b}}$, 16) & 0 & No \\
    \cdashlinelr{1-4}
    Dense & ($n_{\mathrm{b}}$, 1)  &  17 & Output\\
    \bottomrule
    \end{tabular}
    \caption{Critic architecture.}
    \label{tab:gan_archi_crit}
\end{table}

\newpage

 \bibliographystyle{elsarticle-num} 
 \bibliography{cas-refs}





\end{document}